\documentclass{article}

\usepackage[in]{fullpage}  
\usepackage{natbib} 
\usepackage{mathpazo} 
\usepackage[table]{xcolor}
\usepackage[utf8]{inputenc} %
\usepackage[T1]{fontenc}    %
\usepackage{hyperref}       %
\usepackage{url}            %
\usepackage{booktabs}       %
\usepackage{amsfonts}       %
\usepackage{nicefrac}       %
\usepackage{microtype}      %
\usepackage{lipsum}         %
\usepackage{todonotes}
\usepackage{xcolor}

\usepackage{colortbl}

\usepackage{tcolorbox}
\usepackage{xcolor}
\usepackage{framed}
\colorlet{shadecolor}{orange!15}

\usepackage{url}            %
\usepackage{booktabs}       %
\usepackage{multirow}    
\usepackage{amsfonts}       %
\usepackage{nicefrac}       %
\usepackage{microtype}      %
\usepackage{natbib}
\usepackage{enumerate}
\usepackage{hhline}
\usepackage{makecell}
\usepackage{pifont}

\usepackage{graphicx} %
\usepackage{subfigure}
\usepackage{caption}
\usepackage{subcaption}
\usepackage{amsmath}
\usepackage{amsthm}
\usepackage{amssymb}
\usepackage{tikz}
\usepackage{xcolor}
\usetikzlibrary{arrows}

\allowdisplaybreaks

\usepackage{mathrsfs}

\usepackage{algorithm}
\usepackage{algorithmic}
\usepackage{hyperref}
\usepackage{bm}

\allowdisplaybreaks

\usepackage[capitalize,noabbrev]{cleveref}
\crefname{thm}{Theorem}{Theorems}
\crefname{lem}{Lemma}{Lemmas}
\crefname{cor}{Corollary}{Corollaries}
\crefname{prop}{Proposition}{Propositions}
\crefname{asmp}{Assumption}{Assumptions}
\crefname{defn}{Definition}{Definitions}
\crefname{oracle}{Oracle}{Oracles}
\crefname{fact}{Fact}{Facts}
\crefname{conj}{Conjecture}{Conjectures}
\crefname{rem}{Remark}{Remarks}
\crefname{example}{Example}{Examples}
\crefname{condition}{Condition}{Conditions}
\crefname{exercise}{Exercise}{Exercises}
\crefname{algorithm}{Algorithm}{Algorithms}
\crefname{table}{Table}{Tables}
\crefname{figure}{Figure}{Figures}
\crefname{section}{Section}{Sections}
\crefname{subsection}{Section}{Sections}
\crefname{appendix}{Appendix}{Appendices}
\crefname{message}{Message}{Messages}

\definecolor{red}{rgb}{1, 0, 0}

\definecolor{green}{rgb}{0, 1, 0}

\definecolor{blue}{rgb}{0, 0, 1}

\definecolor{orange}{rgb}{1, 0.4, 0.0}

\input{packages/math_commands}

\usepackage{soul}

\usepackage{wrapfig}
\usepackage{ulem} %

\AtBeginDocument{\setlength\abovedisplayskip{5pt}}
\AtBeginDocument{\setlength\belowdisplayskip{5pt}}

\newcommand{\rev}[1]{\textcolor{black}{#1}}
\newcommand{\yupeng}[1]{\textcolor{black}{#1}}

\usepackage{amsmath}   %
\usepackage{booktabs}  %
\usepackage{float}     %
\usepackage{xcolor}    %
\usepackage{graphicx}  %

\usepackage{soul}

\definecolor{revisionred}{RGB}{190,0,0}
\definecolor{deletiongray}{RGB}{85,85,85}
\setstcolor{deletiongray}

\hypersetup{
    colorlinks=true,
    citecolor=blue,
    linkcolor=blue,
}

\usepackage{listings} %

\definecolor{codegreen}{rgb}{0,0.6,0}
\definecolor{codegray}{rgb}{0.5,0.5,0.5}
\definecolor{codepurple}{rgb}{0.58,0,0.82}
\definecolor{codeblue}{rgb}{0,0,1}
\definecolor{backcolour}{rgb}{0.95,0.95,0.92}
\definecolor{key-color}{rgb}{0.8, 0.47, 0.196}

\lstdefinestyle{mystyle}{
    backgroundcolor=\color{backcolour},   
    commentstyle=\color{codegreen},
    numberstyle=\tiny\color{codegray},
    stringstyle=\color{codepurple},
    basicstyle=\ttfamily\footnotesize,
    breakatwhitespace=false,         
    breaklines=true,                 
    captionpos=b,                    
    keepspaces=true,                 
    numbers=left,                    
    numbersep=5pt,                  
    showspaces=false,                
    showstringspaces=false,
    showtabs=false,                  
    tabsize=2,
    language=Python,
    emph={lm},
    emphstyle={\color{blue}},
    classoffset=1, %
    otherkeywords={sum},
    morekeywords={rm, mean},
    keywordstyle=\color{codegreen},
    classoffset=0,
}
\title{TrainBridge: Agentic NPU Training Optimization Guided by GPU Compilation Evidence}

\makeatletter
\def\@fnsymbol#1{\ensuremath{\ifcase#1\or *\or \dagger\or \ddagger\or
   \mathsection\or \sharp\or \Diamond\or \mathparagraph\or \|\or
   \or \ddagger\ddagger \else\@ctrerr\fi}}
\makeatother

\usepackage{authblk}

\author[1,2]{Yupeng Chen\thanks{Equal contribution. Email: \{yupengchen1, kunxiangzhao\}@link.cuhk.edu.cn}}
\author[2]{Kunxiang Zhao\textsuperscript{*}}
\author[2]{Xingjian Wang}
\author[2]{Yue Zhao}
\author[2]{Yijing Wu}
\author[2]{\\ Ye Liu}
\author[3]{Kai Li}
\author[1]{Congliang Chen}
\author[1,2]{Ruoyu Sun\thanks{Correspondence to: Ruoyu Sun (sunruoyu@cuhk.edu.cn).}}

\affil[1]{Shenzhen Loop Area Institute}
\affil[2]{The Chinese University of Hong Kong, Shenzhen, China}
\affil[3]{Autokernel}
\date{}

\begin{document}

\maketitle
\vspace{-8mm}
\begin{abstract}

\rev{We study how to improve training efficiency on Ascend NPUs. In the NPU training programs we study, framework-level compilation optimizations such as operator fusion are only partially applied, leaving fragmented computation and avoidable execution overheads. Profiling-guided tuning can address these inefficiencies, but it often requires costly trial and error.}
\rev{To make training on NPUs more efficient, we propose TrainBridge, an agentic approach for transferring GPU optimization guidance to NPU training.
TrainBridge uses the highly optimized execution produced by mature GPU compilation stacks as a task-specific reference for NPU tuning. It extracts optimization intents from the GPU training execution to guide NPU program rewrites and further refines the rewritten program through iterative profiling.}
\rev{Our GPU-to-NPU migration experiments show that}
TrainBridge achieves an average 1.7× speedup across four workloads covering LLM pretraining, image classification, and vision-language training, and an average speedup of nearly 2.0× across 27 Modded-NanoGPT speedrun records with highly optimized GPU implementations.\footnote{Our code is available at \url{https://github.com/YChen-zzz/NPU_optimal_reference_skills}.}

\end{abstract}

\section{Introduction}

Efficient training on modern accelerators can benefit from framework-level compilation that optimizes computation across operators. \rev{Such compilation is mature on NVIDIA GPU backends. However}, in the Ascend NPU training programs we study, \rev{these optimizations are partially applied by the available compilation stack \citep{yu2026limitations}.}
\rev{As a result, the execution of the training algorithm involves  fragmented computation, redundant operations, and materialized intermediates, leaving space for NPU-specific program tuning.}

Such NPU-specific tuning is often manual and expert-driven. Providing a tuning agent with
task-specific knowledge, such as NPU profiling results
\citep{zhang2025cudaforge,TREX,zhou2025accelerating}, enables it to diagnose bottlenecks and improve the NPU program. However, even with profiling guidance, \yupeng{the agent still needs to iteratively analyze, modify, execute, and validate the program}, making the
optimization process time-consuming under a limited tuning budget. This motivates us to seek
additional guidance that can make the tuning process more efficient.

To provide this additional guidance, we introduce TrainBridge, a two-stage agentic
framework for NPU-specific optimization. We observe that the corresponding GPU
training programs already benefit from substantial optimization by mature compilation stacks.
\rev{The differences between the eager and compiled GPU execution reveal optimization decisions already made for the same workload. 
Therefore, we think the differences are useful guidance on NPU execution.}
In Stage 1, TrainBridge extracts the underlying optimization intents from the \yupeng{differences} and
constructs NPU-specific rewrites when corresponding inefficient computation patterns are present
in the NPU program. \rev{In Stage 2, it iteratively profiles the updated program to identify remaining bottlenecks, addresses them, and updates the profile.}

We evaluate TrainBridge in a GPU-to-NPU migration setting on training workloads spanning LLM pretraining, image classification, and vision-language training.

In addition, we introduce an NPU version of the
Modded-NanoGPT speedrun \citep{modded_nanogpt_2024}, a highly optimized community benchmark for end-to-end
language-model pretraining efficiency under a fixed validation target. We adapt its historical
GPU implementations to Ascend NPUs and release an open
leaderboard\footnote{\url{https://github.com/zhaoyiyidan/modded-nanogpt-npu}} and evaluation
environment to support further NPU training optimization.

Our contributions are summarized as follows:
\begin{itemize}
    \item %
    \rev{We identify compilation-related optimization gaps in NPU training programs. To narrow these gaps, we present TrainBridge, a two-stage agentic approach. TrainBridge first extracts optimization intents from the differences between eager and compiled GPU implementations and uses them to guide NPU program rewrites, then iteratively profiles the rewritten program to address residual bottlenecks.}

    \item \rev{We evaluate TrainBridge across multiple NPU training workloads.} It achieves average
        speedups of 1.7× across four training workloads and nearly 2.0× across 27
        historical speedrun records with highly optimized GPU implementations, relative to the initial
        NPU implementations. On the four additional workloads, it reduces the average NPU-to-GPU
        training-time ratio from 2.8× to 1.5×.
\end{itemize}

\section{NPU Training: Compilation Gaps and Optimization Challenges}
\label{backgroud_and_motivation}

\subsection{Training Workloads and Migration Setting}
\label{workloads}

We study GPU-to-NPU training migration across five workloads: Modded-NanoGPT, CIFAR-10 AirBench\footnote{\url{https://github.com/KellerJordan/cifar10-airbench}},
Qwen3-0.6B training with Hugging Face Trainer, ViT training with TIMM\footnote{\url{https://github.com/huggingface/pytorch-image-models}}, and nanoVLM\footnote{\url{https://github.com/huggingface/nanoVLM}}. These workloads
cover language-model pretraining, image classification, and vision-language training. We evaluate them using one to eight devices.

Modded-NanoGPT builds on NanoGPT\footnote{\url{https://github.com/KellerJordan/Modded-NanoGPT}},
a language-model pretraining codebase widely used for studying the architectural properties of
LLMs \citep{zhang2024transformers} and training algorithms
\citep{Mars2025,zhang2025adam,zhang2025adams}. The original GPU speedrun requires participants to train a language model on eight NVIDIA H100 GPUs and achieve a validation loss below 3.28
\citep{modded_nanogpt_2024}. The speedrun contains the GPT-2 small, GPT-2 Medium, and
optimization tracks. We study the GPT-2 small track, which contains a sequence of
high-performance implementations developed over time. These implementations use different
training and kernel optimizations and therefore provide diverse migration tasks. We use
\emph{record} to denote a valid historical leaderboard implementation that reaches the required
loss target; \textbf{a larger record number indicates a later and faster implementation with a
shorter end-to-end training time.}

For all workloads, we first use Kerminal\footnote{\url{https://kerminal.cn/}} to perform the
initial functional migration from GPU to Ascend 910C NPUs. For Modded-NanoGPT, we preserve
the original training schedule, hyperparameters, and model dynamics, and require the migrated
program to reach validation loss below 3.28 at the same training step. The corresponding correctness criteria for the
other workloads are described in Appendix \ref{other_setting}.

\subsection{Compilation Gaps in NPU Training}

Across the four additional workloads besides Modded-NanoGPT, the initial NPU-to-GPU end-to-end
training-time ratio averages 2.8$\times$.
For Record 50, the compiled GPU implementation takes 128 seconds and the initial NPU
implementation takes 563 seconds, giving a ratio of 4.40$\times$. These measurements reflect
differences in both hardware capability and software optimization. Reported
compute-throughput figures\footnote{\url{https://www.nvidia.com/en-us/data-center/h100/}}
\citep{zuo2025serving} suggest that the Ascend 910C provides roughly 76\% of the H100 SXM's peak
dense BF16/FP16 compute throughput. These end-to-end gaps cannot be fully explained by hardware capability alone and motivate us to investigate software-level optimization opportunities in the migrated NPU programs.

\subsection{Challenges in Agentic Performance Tuning}

We first tried using a generic coding agent with GPT-5.6-SOL in the Codex agent harness to tune the initial NPU program. It produced valid changes, but many candidates failed to improve end-to-end training time. Without task-specific knowledge, the agent often struggles to identify which regions have substantial
optimization potential and what changes are likely to improve training efficiency.

NPU profiling provides useful task-specific guidance by identifying expensive operators and
timeline intervals. However, high runtime alone does not indicate how much a region can be accelerated or which optimization should be applied. The agent therefore needs repeated cycles
of analysis, modification, execution, and validation to discover effective changes. While this
process can improve performance, it can be time-consuming and inefficient under a limited tuning
budget. This motivates complementary guidance that can narrow the optimization search \yupeng{space.}

\begin{table}[h]
\centering
\caption{Training times for GPU and NPU implementations of Modded-NanoGPT Record 50. Ratios use the
    compiled GPU program as the reference.}
\label{tab:portability-gap}
\begin{tabular}{lrr}
\toprule
Implementation & Training time (s) & Relative time \\
\midrule
GPU, compiled & 128 & $1.00\times$ \\
GPU, eager & 338 & $2.64\times$ \\
NPU, initial port & 563 & $4.40\times$ \\
\bottomrule
\end{tabular}
\end{table}

\subsection{GPU Compilation as Optimization Evidence}
GPU training ecosystems combine expert-designed implementations with mature framework-level compilation. For the same workload, differences between eager and compiled GPU execution reveal concrete changes that eliminate, fuse, or reorganize computation. Modded-NanoGPT Record 50 provides a representative example. Table~\ref{tab:portability-gap} shows that, on eight NVIDIA H100 SXM GPUs, framework-level compilation through
\texttt{torch.compile} reduces the end-to-end training time of Record 50 from around 338 seconds
to 128 seconds, demonstrating its contribution to training efficiency.

We next examine whether the computation patterns addressed by the GPU-side compilation stack remain in the migrated NPU implementation. In our evaluated NPU training programs, some corresponding patterns remain despite the available compilation support, leaving fragmented computation and materialized intermediates that provide concrete opportunities for NPU-specific optimization. 

Modded-NanoGPT Record 50
provides one example. Its KeyShift operation shifts half of each attention head’s key features by
one token along the sequence dimension. The original overlapping-slice assignment remains and produces a sequence of \texttt{ViewCopy} operations. Consistent with prior reports of compilation limitations on Ascend \citep{yu2026limitations}, the Ascend NPU compilation stack did not successfully produce a runnable end-to-end compiled program in our migrated training setting, leaving this inefficiency in the executed NPU implementation. The same inefficiency is present in eager GPU execution, whereas the compiled GPU execution eliminates it through functionalization and kernel fusion. Then, the agent extracts the intent of avoiding overlapping view updates and proposes an NPU-specific expression that realizes this intent. Table~\ref{tab:keyshift} reports the performance improvement from the NPU-specific rewrite.

\begin{table}[H]
\centering
\caption{Local NPU latency of  KeyShift operation in Record 50 of Modded-NanoGPT speedrun before
    and after GPU-compilation-evidence-guided agent rewriting. The input shape is  [1, 24576, 6, 128], and
    we use BF16 precision.}
\label{tab:keyshift}
\begin{tabular}{lrr}
\toprule
Implementation & Latency (ms) & Speedup \\
\midrule
Original NPU implementation & 2.604 & $1.00\times$ \\
GPU-compilation-evidence-guided agent rewrite & 0.508 & $5.13\times$ \\
\bottomrule
\end{tabular}
\end{table}

This case illustrates that, when a computation pattern eliminated by mature GPU compilation remains in the NPU program, the underlying optimization intent can guide an effective NPU-specific rewrite and improve the efficiency of the corresponding NPU computation.

\section{TrainBridge}
\subsection{Problem Formulation}

We first formalize how source-side framework-level compilation provides optimization evidence for NPU tuning. In our source GPU setting, this compilation is implemented through PyTorch \texttt{torch.compile}. Such compilation may transform computation graphs and execution schedules, introduce fused regions, modify compiler IR, and generate specialized kernels. For clarity, we use computation-graph changes as a representative view to formulate the optimization problem; TrainBridge itself is not restricted to graph-level transformations. Our goal is to improve the end-to-end training time of the migrated NPU program while preserving its correctness requirements under a limited tuning budget.

At the graph level, let $G_e$ and $G_c$ denote representative computation graphs before and after source-side GPU compilation, respectively. %
From the computation-graph perspective, we summarize %
\yupeng{the} changes as
\[
G_c = \tau_G(G_e),
\]
where $\tau_G$ denotes only the graph-level changes observed during compilation. The difference between $G_e$ and $G_c$ therefore provides evidence of computation that the source stack found beneficial to eliminate, fuse, or reorganize.

Within $G_e$, let $r_e \subseteq G_e$ denote a computation region before GPU compilation, and let $r_c$ denote its compiled counterpart. Suppose an inefficient computation pattern $p$ is present in $r_e$ but is eliminated or reorganized in $r_c$. The difference between $r_e$ and $r_c$ therefore provides evidence about the optimization intent applied to this region.

Migration preserves the high-level computation of the source program but does not automatically reproduce the optimizations introduced by GPU compilation. When the available NPU compilation stack does not eliminate the corresponding inefficiency, the inefficient pattern $p$ may remain in a semantically corresponding NPU region $r_n$. 

\yupeng{In such cases, the agent can use} the eager-to-compiled GPU difference to infer the underlying optimization intent and derive an NPU-specific rewrite that removes or reduces the corresponding inefficiency while preserving the original computation semantics.  Applying such rewrites across applicable regions can improve the training efficiency of the migrated NPU program.

However, GPU evidence cannot address all inefficiencies in the migrated NPU program. Some GPU
transformations may not be beneficial on the NPU because of backend-specific execution
characteristics, and GPU evidence does not expose NPU-specific overheads in host scheduling,
communication, or memory management. These limitations motivate the two stages below.

\subsection{Stage 1: NPU Optimization Guided by GPU Compilation Evidence}

\begin{figure*}
    \centering
    \includegraphics[width=0.95\linewidth]{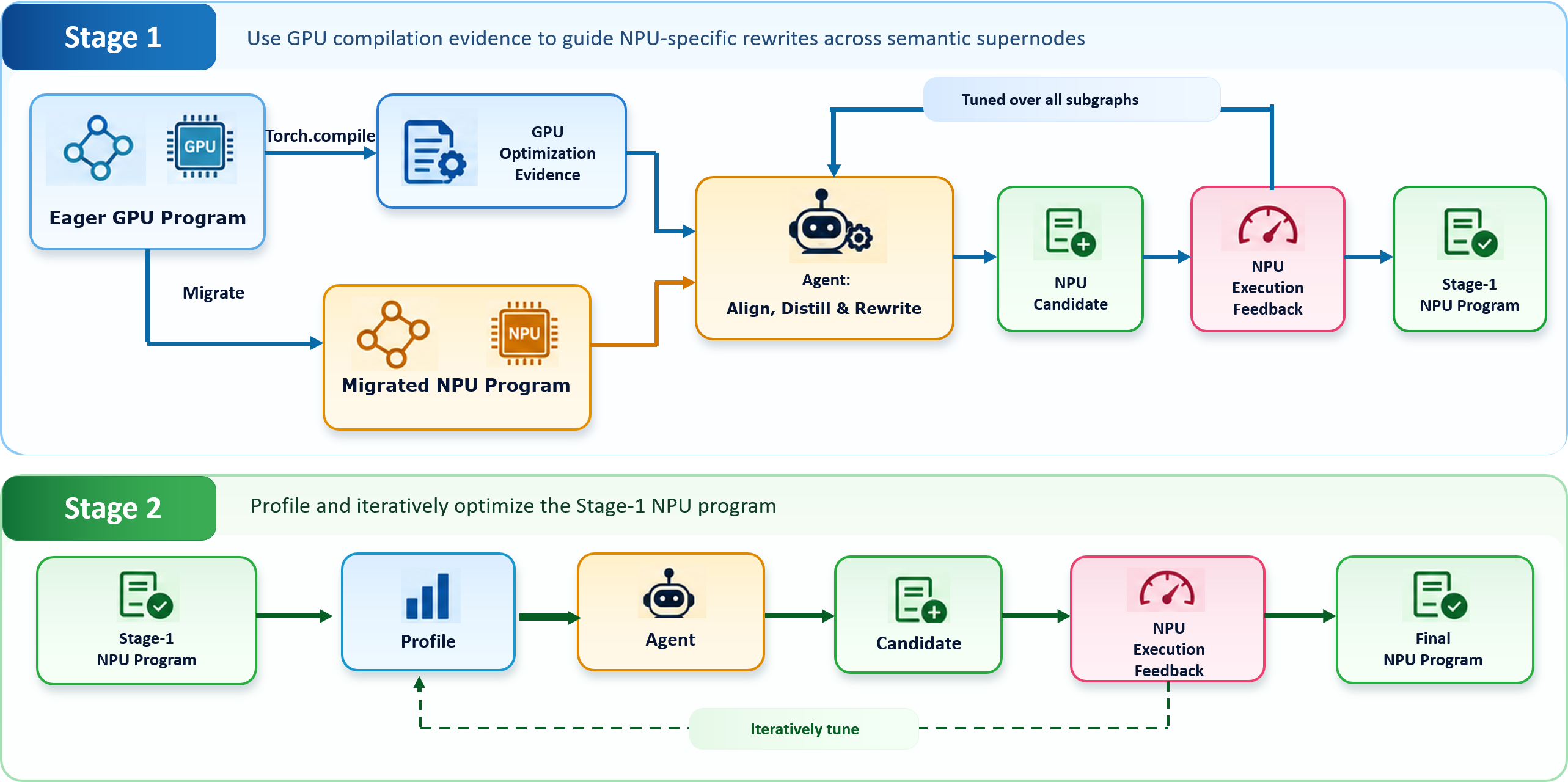}
    \caption{\textbf{Overview of TrainBridge.} Stage 1 aligns eager-to-compiled GPU changes with
        the migrated NPU program, and uses the resulting optimization evidence to guide NPU
        implementation optimization. Stage 2 iteratively profiles the updated NPU program and
        addresses residual bottlenecks.}
\end{figure*}

For the same training program, we collect GPU compilation artifacts, including pre-compilation graphs, post-fusion IR, generated kernels, and profiling summaries, together with the corresponding GPU and migrated NPU sources. Based on this evidence, TrainBridge groups related GPU operations into semantic supernodes. A semantic supernode represents a common high-level computation and provides a unit for relating source-side compilation changes to the migrated NPU program.

For each supernode, the agent first examines the eager-to-compiled GPU changes to identify what computation was eliminated, fused, or reorganized. These changes provide evidence of an inefficient computation pattern in the eager execution and the optimization intent used to address it. The agent then locates the NPU region implementing the same high-level computation and checks whether the corresponding inefficiency remains, using differences in operators, dtypes, layouts, materialized intermediates, and kernel boundaries as evidence.

When the inefficiency remains on the NPU, the agent uses the inferred optimization intent to propose an NPU-specific rewrite that removes or reduces it while preserving the original computation semantics. Candidate rewrites include API parameter changes, redundant-operation removal, fused NPU operators, and selective compilation of supported local subgraphs. Stage 1 ends after all semantic supernodes have been examined and applicable GPU-evidence-guided rewrites have been validated.

\subsection{Stage 2: Profiling-Guided Residual Optimization}
Stage 2 profiles the NPU program produced by Stage 1 to identify residual bottlenecks that are not captured by source-side GPU compilation evidence. The agent
combines operator statistics, host and device timelines, communication events, memory activity,
and source inspection to identify the remaining bottlenecks.

The agent considers several common optimization classes: eliminating redundant work, reusing computed
values or buffers, hiding latency through overlap, and replacing an implementation with an
equivalent NPU-native alternative. After each candidate modification, TrainBridge revalidates
correctness and performance and \yupeng{updates} the profile when the bottleneck shifts. \yupeng{The agent repeats this process until no further performance improvement is observed or the tuning budget is exhausted.}

\section{Experimental Setup}
\label{sec:experimental-setup}

\subsection{Tasks and Migration Protocol}

As described in Section~\ref{workloads}, our evaluation covers five training workloads:
CIFAR-10 AirBench\footnote{\url{https://github.com/KellerJordan/cifar10-airbench}},
Qwen3-0.6B pretraining with Hugging Face Trainer,
ViT training with TIMM\footnote{\url{https://github.com/huggingface/pytorch-image-models}},
nanoVLM\footnote{\url{https://github.com/huggingface/nanoVLM}},
and Modded-NanoGPT.
We migrate each workload to Ascend 910C NPUs and tune the \yupeng{migrated} NPU implementation.
For Modded-NanoGPT, we randomly sample and evaluate 27 historical records from the first 50 leaderboard records,
treating each record as a separate tuning task.
A valid optimization must preserve the original training schedule and reach a validation loss
below 3.28 at the same training step.
The correctness criteria for the other workloads are described in Appendix~\ref{other_setting}.

\subsection{Hardware and Software}

For GPU experiments, we use up to eight NVIDIA H100 SXM GPUs with PyTorch 2.7.0 and CUDA 12.8. For NPU experiments,
we use up to eight Ascend 910C NPUs with PyTorch 2.7.1, CANN 8.5.0, and torch-npu 2.7.1.post2. We report
end-to-end training time, excluding tuning time.
For Modded-NanoGPT, we use eight GPUs and eight NPUs in the respective experiments. Device counts for the other workloads are detailed in Appendix \ref{other_setting}.

\textbf{Agent harness.} Unless otherwise specified, we use the Codex agent harness with GPT-5.6-SOL at medium reasoning effort in our main experiments. We launch most tuning tasks through the Codex desktop application.

\subsection{Research Questions}

All methods start from the same initial NPU implementation. We evaluate TrainBridge through two
research questions:
\begin{itemize}
    \item \textbf{RQ1: Agentic tuning effectiveness.} Does TrainBridge find faster validated NPU
        programs under a fixed tuning budget?
    \item \textbf{RQ2: Stage contribution.} Does GPU-compilation-evidence-guided Stage 1 provide additional benefit beyond profiling-guided Stage 2, and does Stage 2 further improve the program produced by Stage 1?
\end{itemize}

For the controlled Record 50 experiments, we use GPT-5.6-SOL with medium reasoning effort and a
12-hour wall-clock tuning budget. For the RQ1 comparison in Section \ref{compare_ab}, we independently run each method three times. For the RQ2 stage-contribution analysis in Section \ref{ablation_sec}, we perform a single run per method. We evaluate the
best candidate that completes full training and passes the validation-loss criterion.

\begin{figure}[t]
    \centering
    \includegraphics[width=0.4\linewidth]{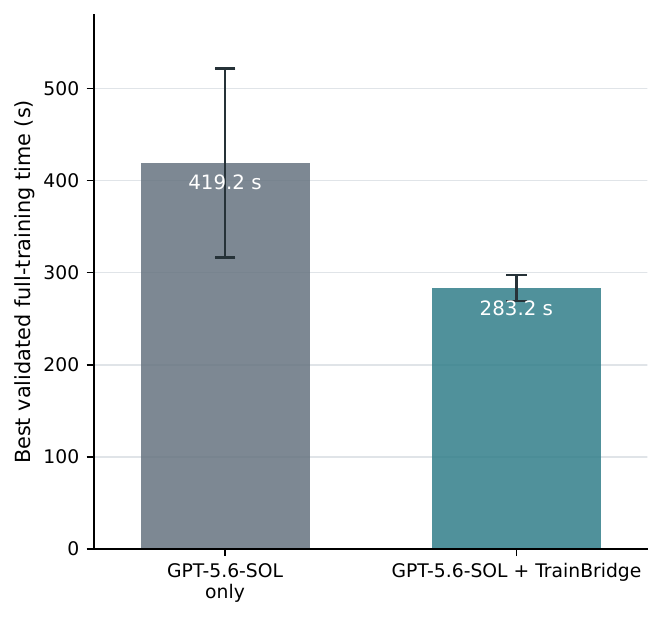}
    \caption{Record 50 of the Modded-NanoGPT speedrun tuning results under a 12-hour budget.
        Results are averaged over three independent runs using the same Codex harness; error
        bars show standard deviation. TrainBridge achieves substantially lower end-to-end
        training time with lower variance across repeated runs.}
    \label{fig:agent-effectiveness}
\end{figure}

\subsection{Overall Agentic Tuning Effectiveness}
\label{compare_ab}

Figure~\ref{fig:agent-effectiveness} compares the generic coding agent with TrainBridge on
Modded-NanoGPT Record 50. Both methods start from the same initial NPU implementation
and use the same 12-hour tuning budget. Across three independent runs, the generic agent
achieves an average best validated training time of 419.2 seconds, whereas TrainBridge
achieves 283.2 seconds, corresponding to a 32.4\% reduction. TrainBridge also exhibits
lower variance across the three runs, indicating more consistent tuning outcomes in this
experiment. Additional evaluation with Claude Opus 4.8 can be seen in Appendix \ref{app:opus}.

\begin{figure}[H]
    \centering
    \includegraphics[width=0.6\linewidth]{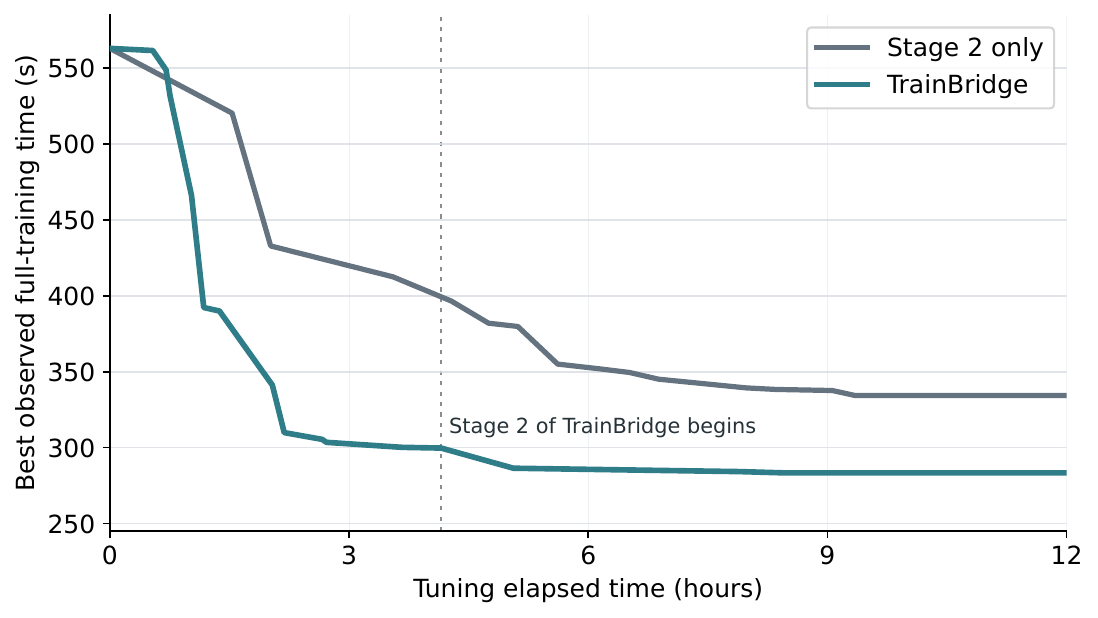}
    \caption{Contribution of the two optimization stages on Record 50 under a 12-hour tuning
        budget. Stage 2 alone uses NPU profiling as task-specific guidance, while TrainBridge
        additionally applies GPU-evidence-guided Stage 1 before profiling-guided Stage 2. GPU
        evidence provides a substantial early gain, and subsequent NPU profiling further
        improves the program.}
    \label{fig:ablation}
\end{figure}

\subsection{Contribution of the Two Stages}
\label{ablation_sec}

To answer RQ2, we compare TrainBridge with an agent that already has task-specific knowledge from
NPU profiling but does not use GPU compilation evidence. Specifically, we use profiling-guided
Stage 2 alone as the baseline and run both methods under the same 
experimental setting as in Section \ref{compare_ab}. As shown in Figure \ref{fig:ablation}, Stage 1
of TrainBridge reduces training time from 563.0 to 299.7 seconds after around 4 hours and Stage 2 further reduces it to 285 seconds. In contrast, Stage 2 alone reaches around 334
seconds and does not fall below 300 seconds within the budget. These results show that GPU compilation evidence provides substantial additional guidance beyond profiling alone, while the further improvement after Stage 1 shows that target-side profiling remains complementary.

\subsection{Results across Modded-NanoGPT Records}

\begin{figure}[h]
    \centering
    \includegraphics[width=0.8\linewidth]{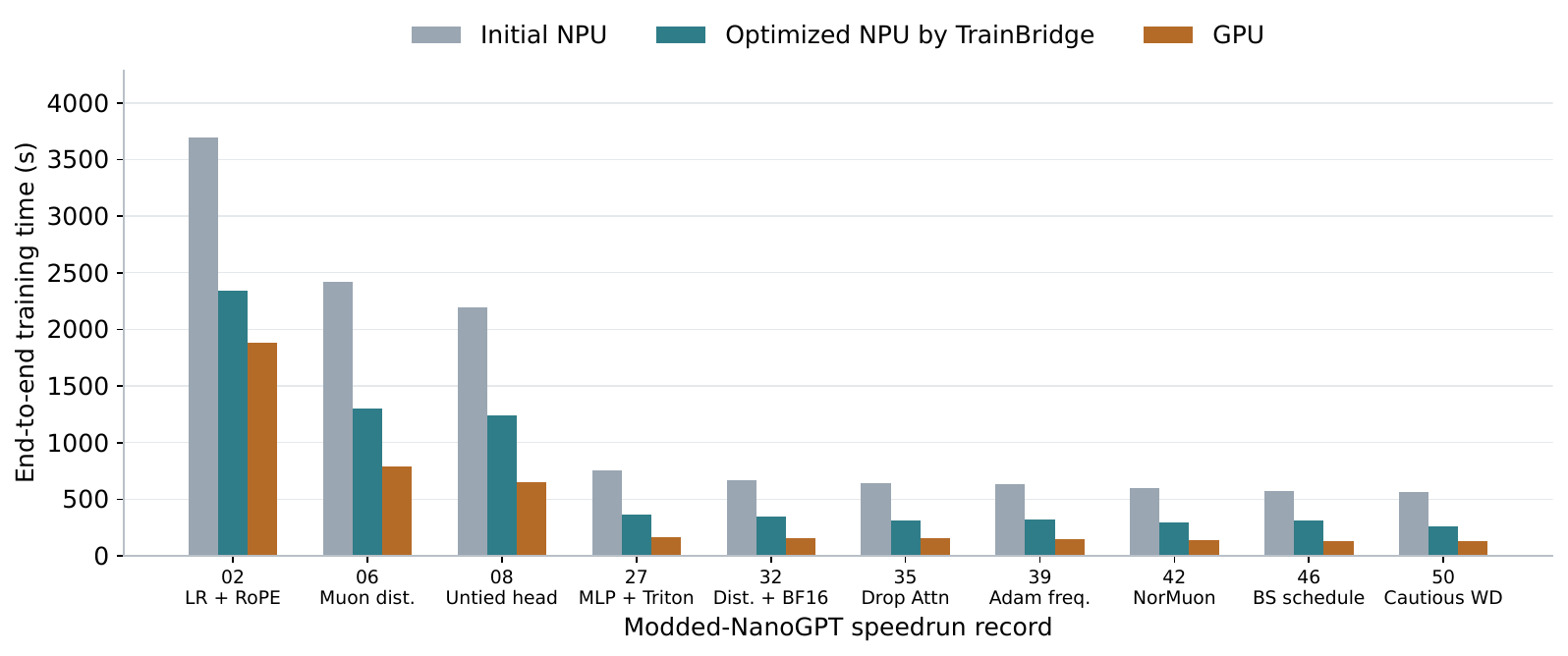}

     \includegraphics[width=0.8\linewidth]{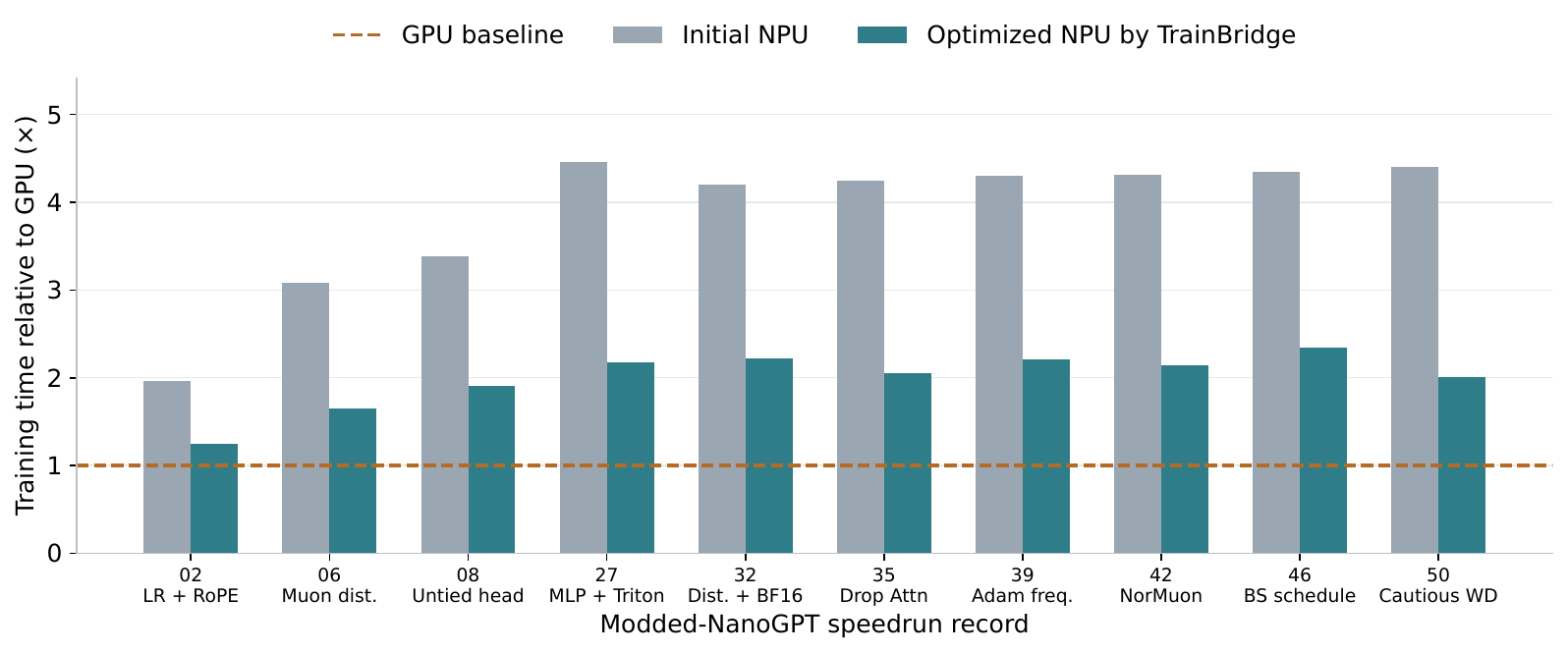}
    \caption{End-to-end training time (top) and training time relative to the GPU baseline
        (bottom) for ten Modded-NanoGPT records sampled approximately uniformly across the
27 evaluated records when ordered by leaderboard record number. TrainBridge consistently reduces
        NPU training time across the shown records; complete results for all 27 evaluated
        records are reported in Appendix \ref{app:nanogpt-results}.}
    \label{fig:nanogpt-summary}
\end{figure}

For each record, we run TrainBridge once with a tuning budget of up to 18 hours and report the best validated training time found within this budget. Therefore, these results may be distinct from the controlled 12-hour experiments in Sections \ref{compare_ab} and \ref{ablation_sec}. Across the 27 evaluated records, TrainBridge achieves an average \(1.95\times\) speedup over the
initial NPU implementations, reducing the average NPU/GPU training-time ratio from \(4.0\times\)
to \(2.0\times\). For readability, Figure~\ref{fig:nanogpt-summary} shows ten records sampled approximately uniformly across the 27 \yupeng{tuned} records when ordered by leaderboard record number. The complete results for all 27 records are reported in Appendix~\ref{app:nanogpt-results}. All optimized
programs satisfy the validation-loss requirement, \yupeng{and results show} that the improvement extends beyond
the Record 50 case study.

\begin{figure}[h]
    \centering
    \includegraphics[width=0.5\linewidth]{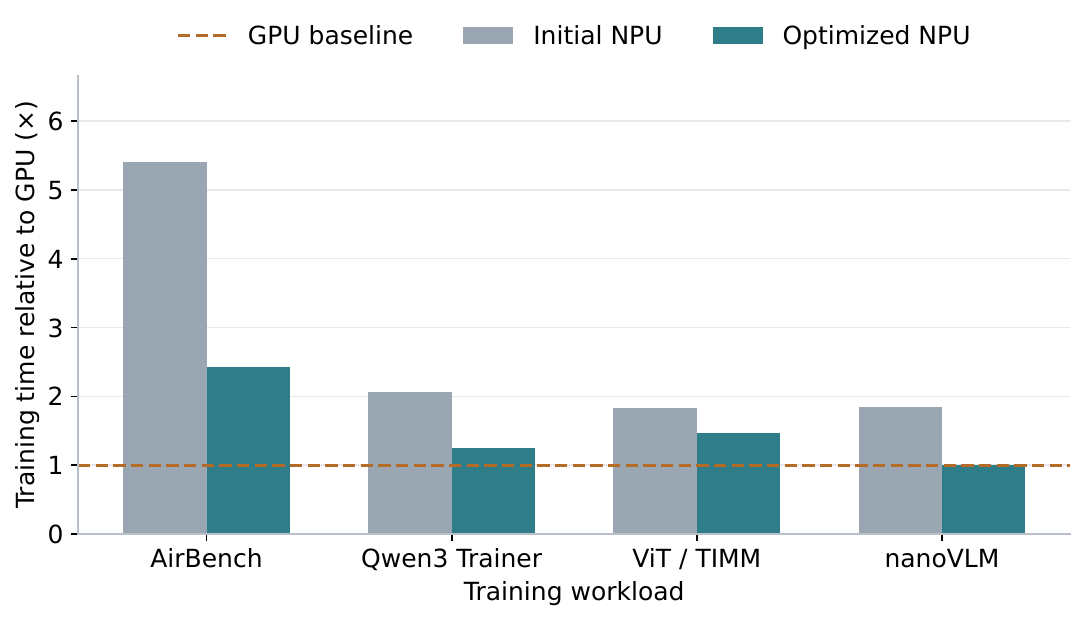}
    \caption{Training time relative to the compiled GPU baseline across four additional
        workloads. Lower is better; the dashed line indicates equal training time on GPU and
        NPU.}
    \label{fig:generalization_to_other_workloads}
\end{figure}

\subsection{Results on Additional Training Workloads}

For each workload, we run TrainBridge once with an 8-hour tuning budget and report the best validated implementation found within the budget. Figure~\ref{fig:generalization_to_other_workloads} compares four additional workloads: CIFAR-10
AirBench, Qwen3 training with Hugging Face Trainer, ViT training with TIMM, and nanoVLM. They
cover image classification, language-model training, and vision-language training. TrainBridge achieves speedups of 2.22×, 1.68×, 1.25×, and 1.84× over the corresponding initial NPU implementations, respectively. These results show that the benefits of TrainBridge extend beyond Modded-NanoGPT to different training workloads.

\section{Related Works}

\paragraph{Agentic performance optimization.}
LLM agents have been used to generate and optimize GPU kernels through execution feedback,
hardware profiling, iterative search, and reusable optimization experience
\citep{dai2026cuda,zhang2025cudaforge,dong2026stark,sun2026kernelskill}. KernelBench provides a
common benchmark for kernel correctness and speed \citep{kernelbench}. Recent work extends
agentic optimization to NPU kernel synthesis and broader ML systems, including end-to-end LLM
deployment on AMD NPUs \citep{li2026human}, Ascend C kernel optimization in large-scale training
\citep{TREX}, and GPU training-engine construction \citep{he2026forgetrain}.

\paragraph{Guiding the optimization search.}
At each iteration, an optimization agent must decide where to intervene and which change
to try. Existing methods improve these decisions in two ways. \yupeng{Training-based methods improve the underlying LLM through task-specific training}; CUDA Agent uses large-scale agentic reinforcement learning with execution-based
rewards to learn CUDA optimization behavior \citep{dai2026cuda}. Training-free methods instead
augment inference-time guidance: CUDAForge uses hardware profiles to diagnose bottlenecks
\citep{zhang2025cudaforge}, STARK organizes prior candidates in a search tree to guide planning
\citep{dong2026stark}, and KernelSkill retrieves reusable expert knowledge while tracking the
current optimization trajectory \citep{sun2026kernelskill}. EvoKernel similarly retrieves
value-weighted experience for NPU kernel synthesis \citep{zheng2026towards}.

TrainBridge additionally uses eager-to-compiled differences from the same GPU workload as
task-specific guidance for where to tune and what to try on the NPU. Target-side profiling then
exposes bottlenecks beyond the GPU compilation evidence.

\section{Limitations and Future Work}
\label{sec:limitations}

Our current tuning setting focuses on single-node training. Multi-node training introduces
additional \yupeng{performance factors} from network topology, cross-node communication, and parallelization
strategy. Future work will evaluate TrainBridge in multi-node settings using large-scale training frameworks such as MindSpeed-LLM and Megatron-LM.

TrainBridge requires a source GPU implementation from which useful framework-level compilation evidence can be obtained. This
requirement provides task-specific \yupeng{optimization guidance} but limits the method to workloads supported by a
mature source stack. A natural extension is to store source-side optimization evidence produced by the GPU compilation stack, together with validated NPU implementations, in a retrieval database.
\yupeng{Such a database could provide reusable knowledge about source-side optimization intents and their successful NPU implementations}, helping agents prioritize promising rewrites for new workloads.

\section{Conclusion}

We study compilation-related optimization \yupeng{inefficiencies} in NPU training and investigate how \yupeng{evidence from mature GPU compilation stacks} can guide NPU-specific program optimization. TrainBridge extracts
optimization intents from \yupeng{differences between eager and compiled GPU implementations,} constructs NPU-specific rewrites,
and profiles the updated program to address residual bottlenecks. It achieves a $2.0\times$ NPU speedup
across Modded-NanoGPT speedrun records and a $1.7\times$ speedup across four additional
workloads. These results show that GPU compilation evidence can guide NPU-specific program
optimization and help address compilation-related inefficiencies in NPU training.

\bibliographystyle{abbrvnat}
\bibliography{reference.bib}

\appendix
\onecolumn

\section{Workloads for other settings}

\subsection{Additional Training Workloads}
\label{other_setting}
We evaluate four additional GPU-to-NPU training workloads. For Qwen3 and nanoVLM, we require the optimized NPU program's final validation
loss to differ from that of the initial NPU port by at most 0.01. For ViT, we apply the same threshold to the final training loss.

\paragraph{AirBench.}
CIFAR-10 AirBench provides training speedruns targeting 94\% and 96\%
accuracy.\footnote{\url{https://github.com/KellerJordan/cifar10-airbench}}
We use \texttt{airbench96\_fast}, which the repository reports reaches
96\% accuracy in under 30 seconds on an A100. We remeasure its GPU
training time on one H100 and evaluate the NPU port on one Ascend 910C.
The optimized NPU program must achieve mean accuracy of at least 96\%.

\paragraph{Qwen3 with Hugging Face
    Trainer.}\footnote{https://github.com/huggingface/transformers/tree/main/examples/pytorch/language-modeling}
We train a randomly initialized Qwen3-0.6B on SlimPajama-6B for 2,000
steps, using a sequence length of 4,096 and a global batch size of 64.
We use eight H100 GPUs and eight Ascend 910C NPUs for the GPU and NPU runs, respectively. We compare validation loss after the final step.

\paragraph{ViT with TIMM.}
Using TIMM,\footnote{\url{https://github.com/huggingface/pytorch-image-models}}
we train a randomly initialized ViT-Base on CIFAR-10 for 200 epochs.
We use four H100 GPUs and four Ascend 910C NPUs for the GPU and NPU runs, respectively.
We compare end-to-end training time.

\paragraph{nanoVLM.}
We migrate the default nanoVLM\footnote{https://github.com/huggingface/nanoVLM} training
configuration.
We use eight H100 GPUs and eight Ascend 910C NPUs for the GPU and NPU runs, respectively. We perform 10 warmup steps before measuring training time over 200 optimizer steps.

\newpage

\section{Additional Evaluation with Claude Opus 4.8}
\label{app:opus}

We additionally evaluate whether the benefit of TrainBridge extends to a different
agent model and harness. Following the Record 50 setting in Section \ref{compare_ab}, we compare
Claude Opus 4.8 running in Claude Code CLI at medium reasoning effort with the same agent augmented by TrainBridge.
Both methods start from the same initial NPU implementation and use the same
execution and validation interfaces.

\begin{figure}[h]
    \centering
    \includegraphics[width=0.4\linewidth]{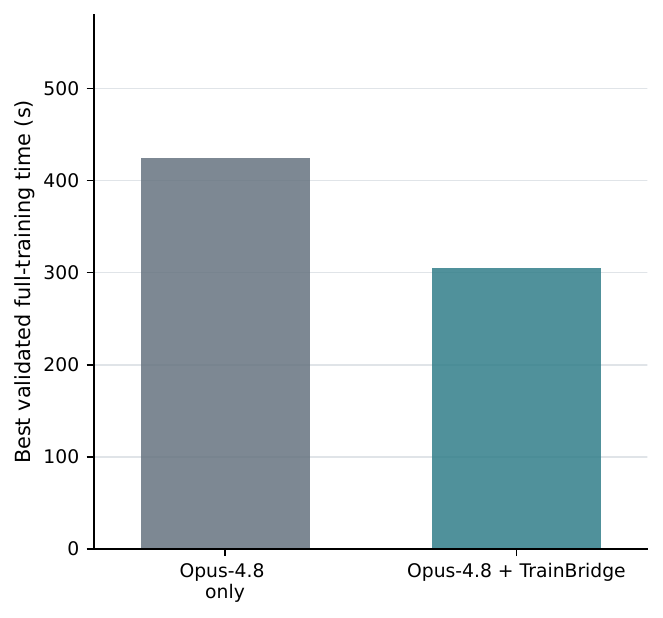}
    \caption{Record 50 of the Modded-NanoGPT speedrun tuning results under an 8-hour budget using Claude Opus 4.8 with the Claude Code harness. TrainBridge achieves substantially lower end-to-end training time than Claude Opus 4.8 alone.}
    \label{fig:agent-effectiveness_cc}
\end{figure}

Due to time and computational-resource constraints, we limit this experiment to an
8-hour tuning budget and perform a single run for each method. The result can be seen in Figure \ref{fig:agent-effectiveness_cc}. Within the 8-hour budget, Claude Opus 4.8 with TrainBridge finds a faster validated
NPU implementation than Claude Opus 4.8 alone. This result is consistent with the
main experiment in Section \ref{compare_ab} and suggests that GPU-compilation evidence can
provide useful optimization guidance beyond the agent model used in our primary
evaluation.

\newpage

\section{Per-Record Modded-NanoGPT Results}
\label{app:nanogpt-results}

Table~\ref{tab:all-nanogpt-records} reports the per-record training times
obtained during tuning. The averages include all 27 records with complete
measurements and are computed over the unrounded per-record ratios, while the average NPU speedup is reported to two decimal places.

\begin{table}[htbp]
\centering
\caption{Modded-NanoGPT per-record results. Times are end-to-end training
times in seconds. NPU speedup is initial NPU time divided by optimized NPU
time. Ratios and
averages are computed before rounding to one decimal place.  All 27 records with complete times
satisfy the validation-loss requirement.}
\label{tab:all-nanogpt-records}
\small
\setlength{\tabcolsep}{4pt}
\begin{tabular}{r r r r r c}
\toprule
Record & GPU (s) & \shortstack{Initial NPU\\(s)} & \shortstack{Tuned NPU\\(s)} &
\shortstack{NPU\\speedup} & \shortstack{NPU/GPU\\before $\rightarrow$ after} \\
50 & 128 & 563 & 257 & $2.2\times$ & $4.4\times \rightarrow 2.0\times$ \\
49 & 129 & 569 & 278 & $2.0\times$ & $4.4\times \rightarrow 2.2\times$ \\
48 & 130 & 574 & 277 & $2.1\times$ & $4.4\times \rightarrow 2.1\times$ \\
47 & 132 & 542 & 278 & $2.0\times$ & $4.1\times \rightarrow 2.1\times$ \\
46 & 132 & 574 & 309 & $1.9\times$ & $4.4\times \rightarrow 2.3\times$ \\
45 & 134 & 579 & 293 & $2.0\times$ & $4.3\times \rightarrow 2.2\times$ \\
42 & 139 & 599 & 298 & $2.0\times$ & $4.3\times \rightarrow 2.1\times$ \\
41 & 141 & 581 & 313 & $1.9\times$ & $4.1\times \rightarrow 2.2\times$ \\
40 & 142 & 610 & 296 & $2.1\times$ & $4.3\times \rightarrow 2.1\times$ \\
39 & 147 & 632 & 325 & $1.9\times$ & $4.3\times \rightarrow 2.2\times$ \\
38 & 149 & 646 & 312 & $2.1\times$ & $4.3\times \rightarrow 2.1\times$ \\
37 & 149 & 560 & 313 & $1.8\times$ & $3.8\times \rightarrow 2.1\times$ \\
35 & 152 & 646 & 312 & $2.1\times$ & $4.3\times \rightarrow 2.1\times$ \\
34 & 153 & 663 & 318 & $2.1\times$ & $4.3\times \rightarrow 2.1\times$ \\
33 & 154 & 655 & 325 & $2.0\times$ & $4.3\times \rightarrow 2.1\times$ \\
32 & 158 & 664 & 341 & $1.9\times$ & $4.2\times \rightarrow 2.2\times$ \\
30 & 162 & 699 & 384 & $1.8\times$ & $4.3\times \rightarrow 2.4\times$ \\
27 & 169 & 755 & 367 & $2.1\times$ & $4.5\times \rightarrow 2.2\times$ \\
17 & 214 & 1183 & 458 & $2.6\times$ & $5.5\times \rightarrow 2.1\times$ \\
10 & 468 & 2038 & 962 & $2.1\times$ & $4.4\times \rightarrow 2.1\times$ \\
09 & 492 & 1910 & 1051 & $1.8\times$ & $3.9\times \rightarrow 2.1\times$ \\
08 & 648 & 2196 & 1230 & $1.8\times$ & $3.4\times \rightarrow 1.9\times$ \\
07 & 720 & 2407 & 1312 & $1.8\times$ & $3.3\times \rightarrow 1.8\times$ \\
06 & 786 & 2420 & 1295 & $1.9\times$ & $3.1\times \rightarrow 1.6\times$ \\
05 & 912 & 2531 & 1357 & $1.9\times$ & $2.8\times \rightarrow 1.5\times$ \\
04 & 1338 & 2426 & 1900 & $1.3\times$ & $1.8\times \rightarrow 1.4\times$ \\
02 & 1884 & 3696 & 2337 & $1.6\times$ & $2.0\times \rightarrow 1.2\times$ \\
\midrule
Average (27 records) & -- & -- & -- & $1.95\times$ &
$4.0\times \rightarrow 2.0\times$ \\
\bottomrule
\end{tabular}
\end{table}

\end{document}